\documentclass[11pt]{article}
\usepackage{psfig,here,fullpage}

\title{Annotation Style Guide \\ for the \\ Blinker Project \\
\vspace*{.4in} Version 1.0.4 \vspace*{.5in}}
\author{I. Dan Melamed \\ 
Dept. of Computer and Information Science \\
University of Pennsylvania \\ 
Philadelphia, PA, 19104, U.S.A. \\
{\tt melamed@unagi.cis.upenn.edu}}

\newcommand{\ignore}[1]{}
\begin{document}
\maketitle

\tableofcontents

\newpage

\section{About This Guide}

This annotation style guide was created by and for the Blinker project
at the University of Pennsylvania.  The Blinker project was so named
after the ``bilingual linker'' GUI, which was created to enable
bilingual annotators to ``link'' word tokens that are mutual
translations in parallel texts.  The parallel text chosen for this
project was the Bible, because it is probably the easiest text to
obtain in electronic form in multiple languages.  The languages
involved were English and French, because, of the languages with which
the project co-ordinator was familiar, these were the two for which a
sufficient number of annotators was likely to be found.

The style guide was created as follows:
\begin{enumerate}
\item The project
co-ordinator wrote a draft version of the General Guidelines in
Section~\ref{gen}.  
\item Two groups of annotators each annotated a set of ten randomly
selected verse pairs from the Bible bitext, using the General
Guidelines draft.  There were nine annotators, so one set of ten verse
pairs was annotated four times and the other five times.
\item The different annotations for each verse pair set were
automatically compared to find differences.
\item The project
co-ordinator manually sorted the sources of variation into about
12 categories.
\item Four of the 9 annotators were reconvened, and presented with
examples of the different types of inter-annotator variation, one type
of variation at a time.  For each kind of variation, there was a brief
discussion, and then a vote took place on the preferred annotation
style.
\item The project co-ordinator compiled the votes and the
examples on which they were based into the Detailed Guidelines in
Section~\ref{det}.  Some clarifying examples were also added to
Section~\ref{det} post-hoc.  
\item As the annotation project got into full swing, annotators
reported a few additional difficult cases.  The project co-ordinator
emailed the problems to all annotators and collected their votes on
the preferred annotation style.  The majority opinions were
incorporated into new versions of the style guide.
\end{enumerate}

\section{General Guidelines}
\label{gen}

You will be working with pairs of corresponding Bible verses in
English and French.  Your task will be to specify how words correspond
within the paired verses, using the Blinker.  For example, when the
Blinker presents you with the pair of verses in Example~1, you might
link them as in Example~2.  As you can see, most words are linked to
only one word in the other language.  However, this is not always the
case, as demonstrated by ``toute'' and ``leur'' in this example.

Sometimes you will see the English on the left and the French on the
right, sometimes vice versa.  You will also notice that we have done
some ``retokenization'' on some of the verses.  In both the English
and the French, we separate hyphenated words and elisions into
separate words.  For example, you will see ``de le'' instead of ``du''
in French, and ``Lord's'' will appear as ``Lord 's'' in English.
Although this is an unusual way of writing, it will make it easier for
you to link the words correctly.

\newpage
Two kinds of complications arise when the translation is not very
literal.

\subsection{Omissions in Translation}

You may see words in the verse of one language whose
meaning is not contained at all in the verse of the other language.
Here is another verse pair from Genesis:

\vspace{.1in}
\noindent French: {\em fixe moi ton salaire , et je te le donnerai .}

\noindent English: {\em And he said , Appoint me thy wages , and I will
give it .}

\vspace{.1in}
Although the English verse begins ``And he said,'' there is no
corresponding language in the French verse.  When this happens, you
should link the extraneous words to the ``Not Translated'' bar on
the corresponding side of the screen, like in Example 3.

Careful!  Many of the translations are very non-literal.  However, you
should only link words to ``Not Translated'' when you can answer
``Yes'' to the following question: If the seemingly extraneous words
were simply deleted from their verse, would the two verses become more
similar in meaning?  If the answer is ``No'' then some words in the
translation share some meaning with some of the words that seem
extraneous.  So, those words are not really extraneous and should not
be marked ``Not Translated.''

\subsection{Phrasal Correspondence}

The other problem with non-literal translations is that sometimes it
is necessary to link entire phrases to each other.  Here is
another example from Genesis:

\vspace{.1in}
\noindent English: {\em And Noah began to be an husbandman , and he planted
a vineyard :}

\noindent French: {\em No\`{a} commen\c{c}a \'{a} cultiver la terre ,
et planta de la vigne.}

\vspace{.1in}
The words in ``to be a husbandman'' and in ``cultiver la terre'' do not
correspond one-to-one, although the two phrases mean the same thing in
this context.  Therefore, the two phrases should be linked as wholes,
by linking each word in one to each word in the other, like in Example
4.  Likewise, ``de la vigne'' means ``some vines,'' not ``a
vineyard.''  Example 4 shows these phrases as completely interlinked.

The divergence in meaning may be so great for some pairs of passages,
that it may seem like the best annotation would be to link both
passages to ``Not Translated'' in their entirety.  Whenever you have
this urge, please remember that neither version of the Bible from
which we drew these verses is a translation of the other.  Instead,
they are both translations of a third version.  Each translation
introduces some idiosyncrasies, and when two such idiosyncrasies
happen in the same place in the text, the two passages may seem like
they have nothing to do with each other.  The decision whether to link
or not to link should {\em not} be based on the question of whether
one passage could have arisen as a translation of the other.  A more
appropriate question is:  Could both of the passages have arisen as
translations of a third.

\begin{figure}[H]
\centerline{\psfig{figure=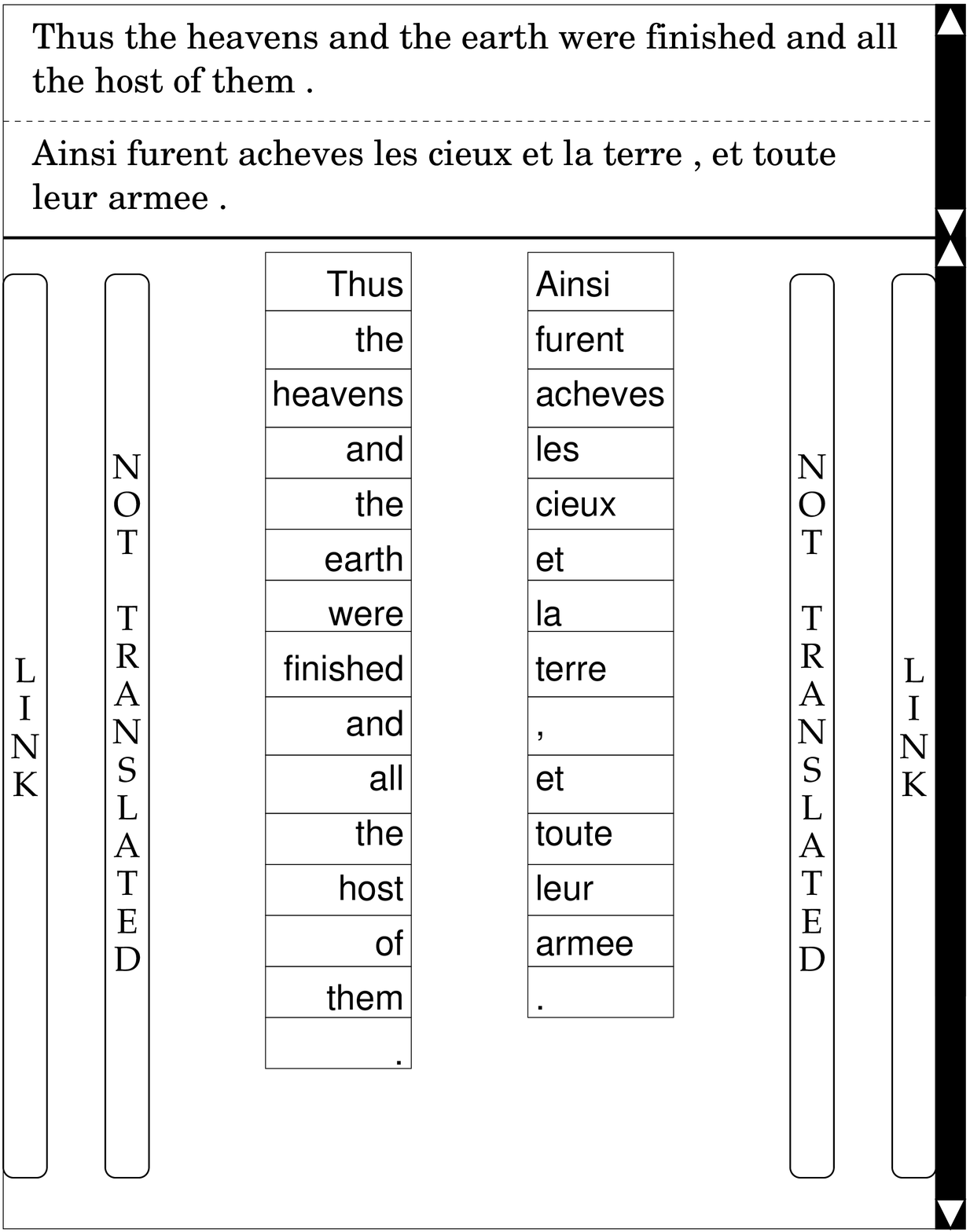,width=6.5in}}
\caption{{\em Example 1.}}
\label{eg0}
\end{figure}

\begin{figure}[H]
\centerline{\psfig{figure=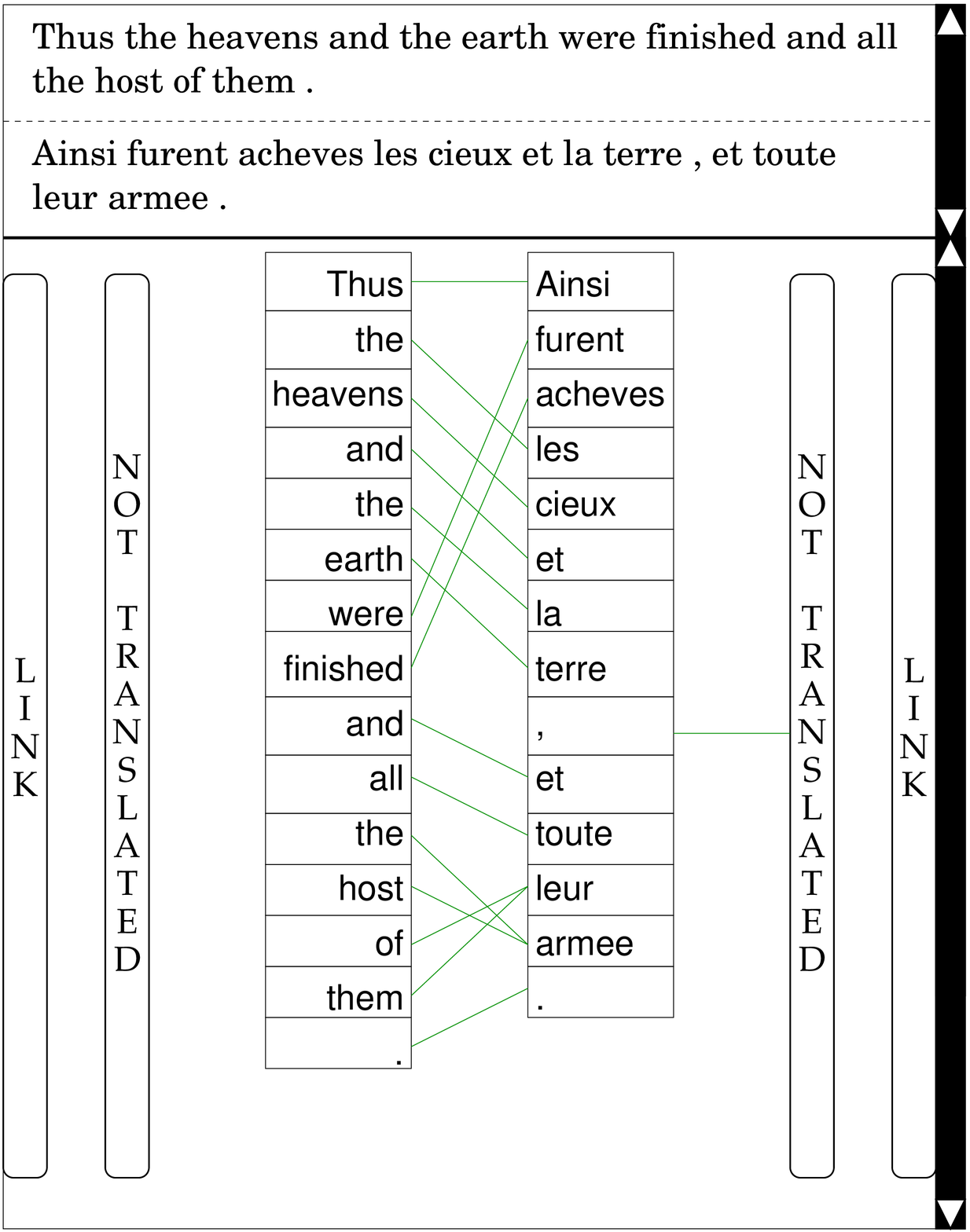,width=7in}}
\caption{{\em Example 2.}}
\label{eg1}
\end{figure}

\begin{figure}[H]
\centerline{\psfig{figure=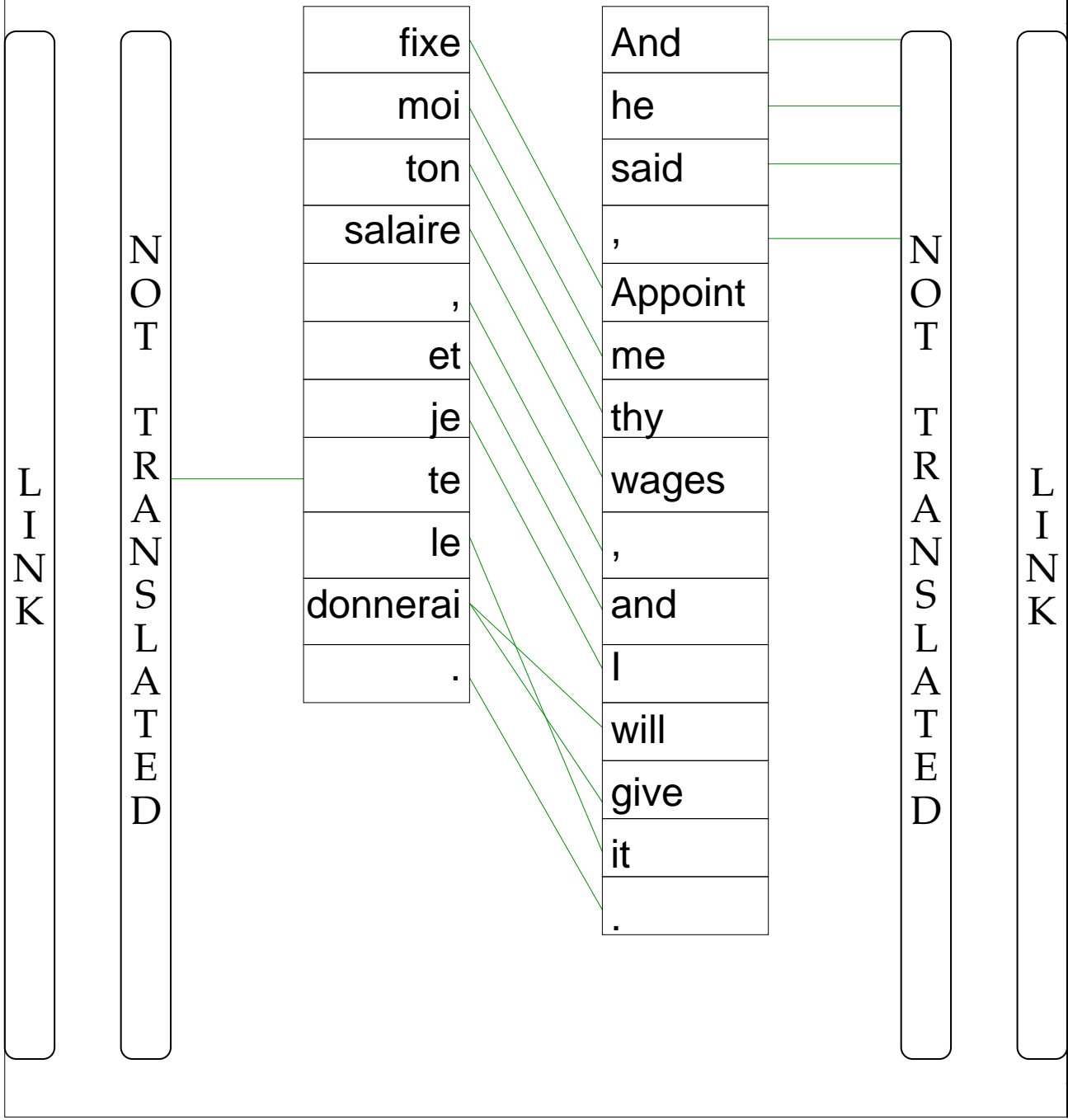,width=7in}}
\caption{{\em Example 3.}}
\label{eg2}
\end{figure}

\begin{figure}[H]
\centerline{\psfig{figure=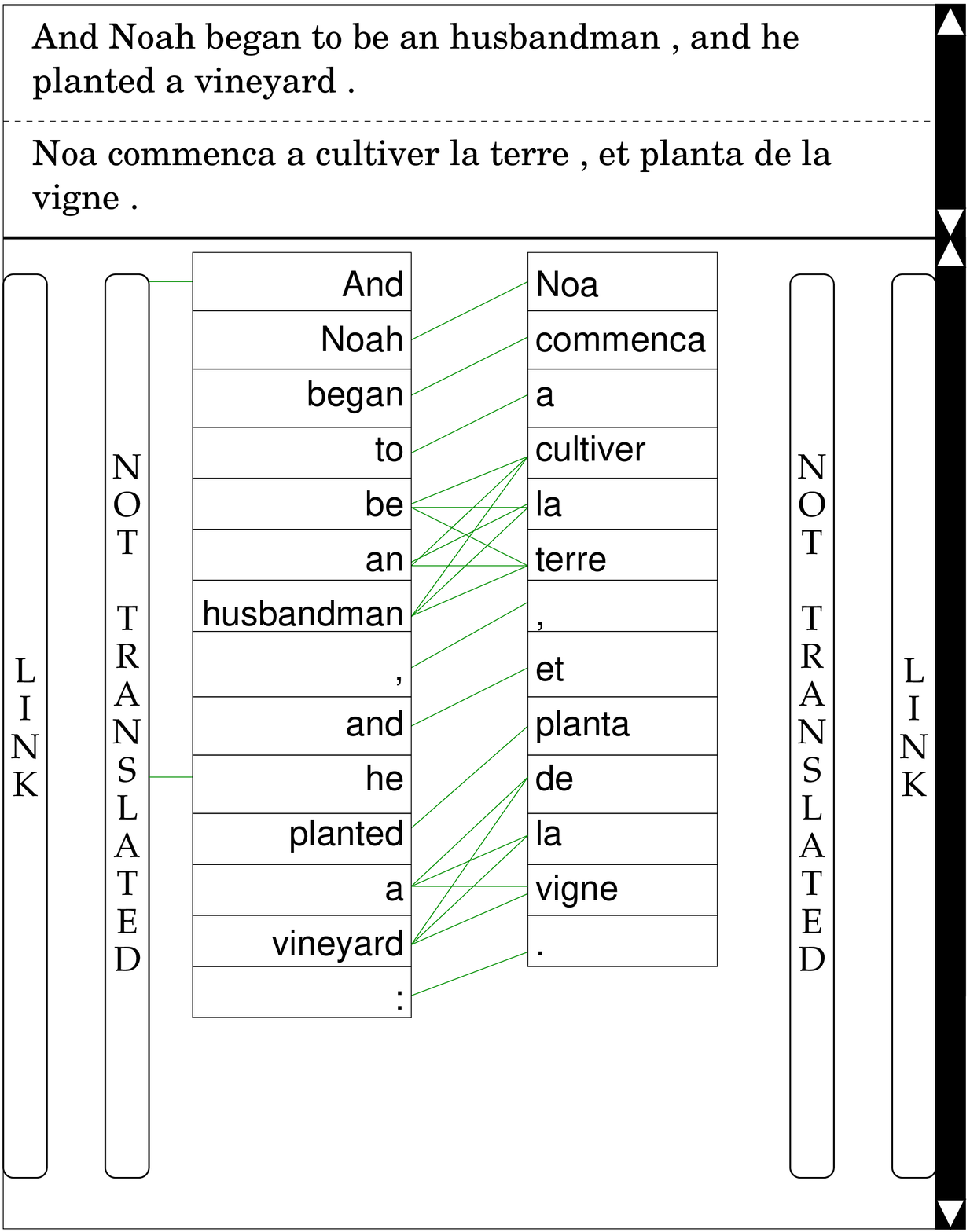,width=7in}}
\caption{{\em Example 4.}}
\label{eg3}
\end{figure}

\newpage

\section{Detailed Guidelines}
\label{det}

You should specify as detailed a correspondence as possible, even when
non-literal translations make it difficult to find corresponding
words.  Here are some examples:

\vspace*{1in}

Right:

\centerline{\psfig{figure=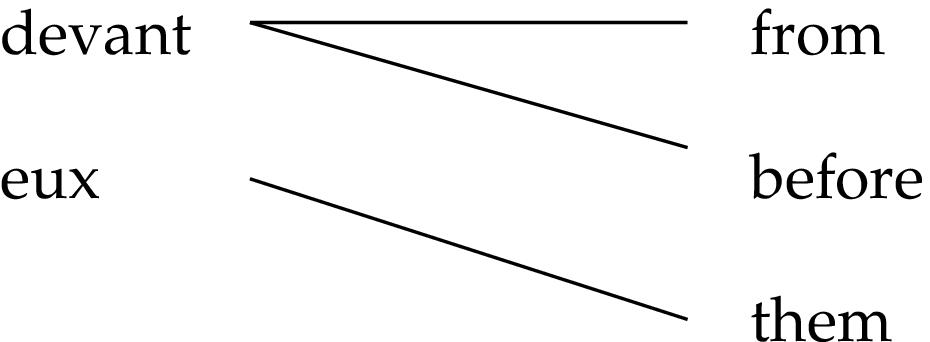,width=3in}}

Wrong:

\centerline{\psfig{figure=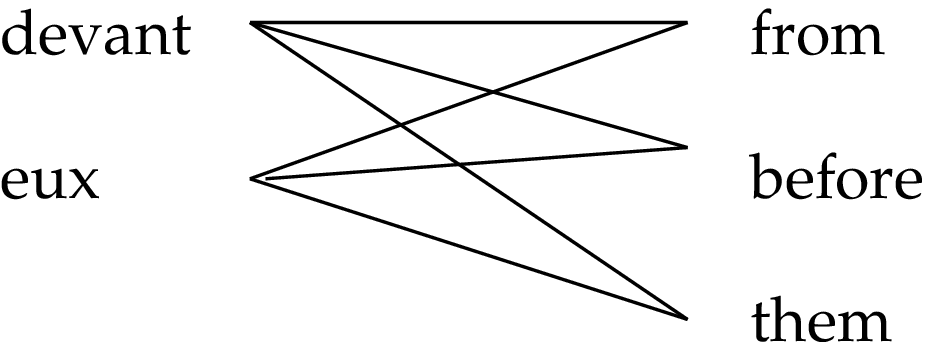,width=3in}}

Right:

\centerline{\psfig{figure=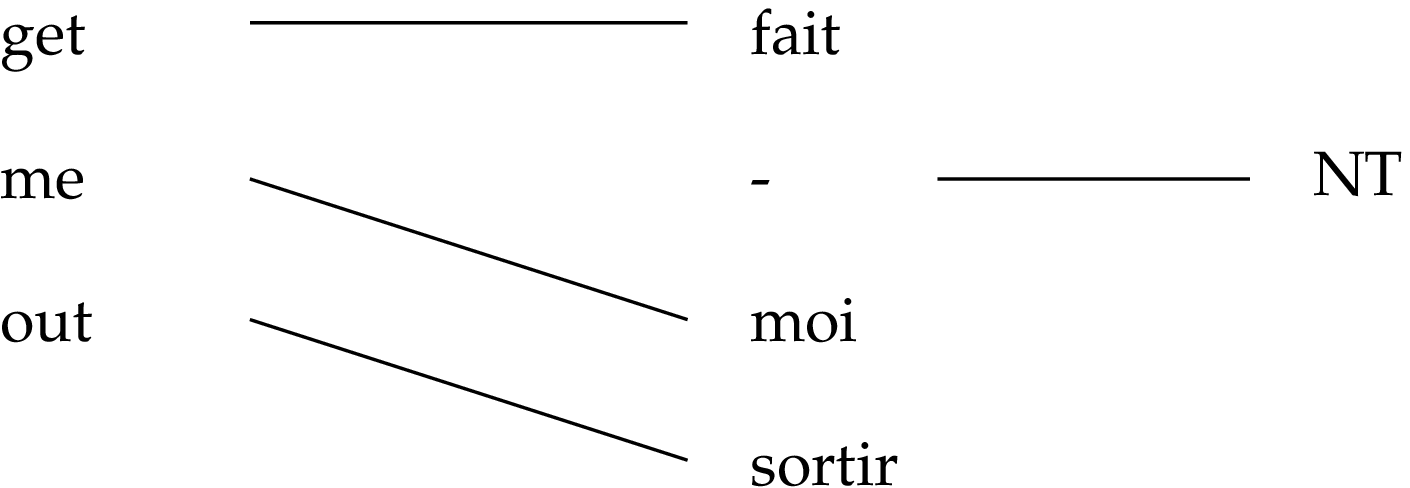,width=4.5in}}

Wrong:

\centerline{\psfig{figure=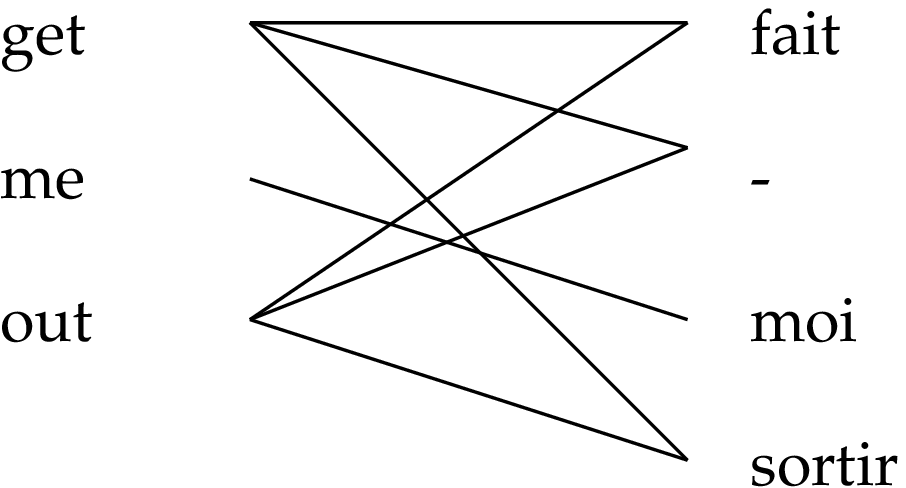,width=3in}}

\newpage

Right:

\centerline{\psfig{figure=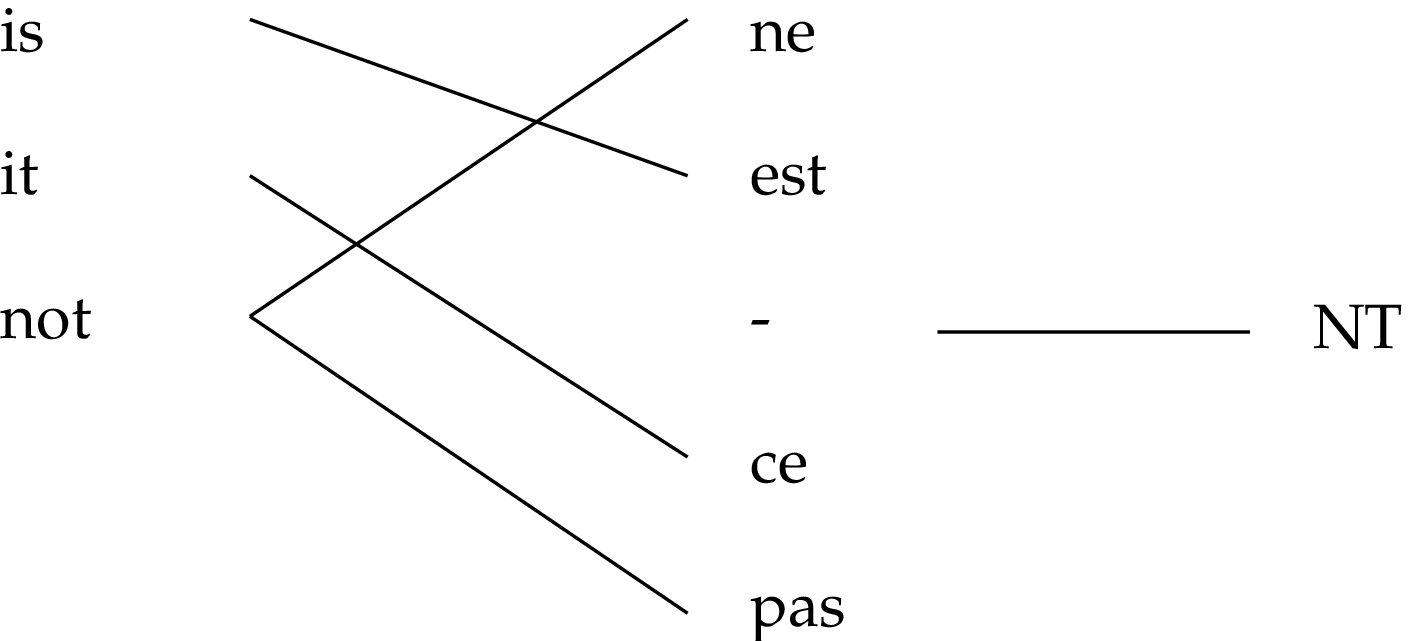,width=4.5in}}

Wrong:

\centerline{\psfig{figure=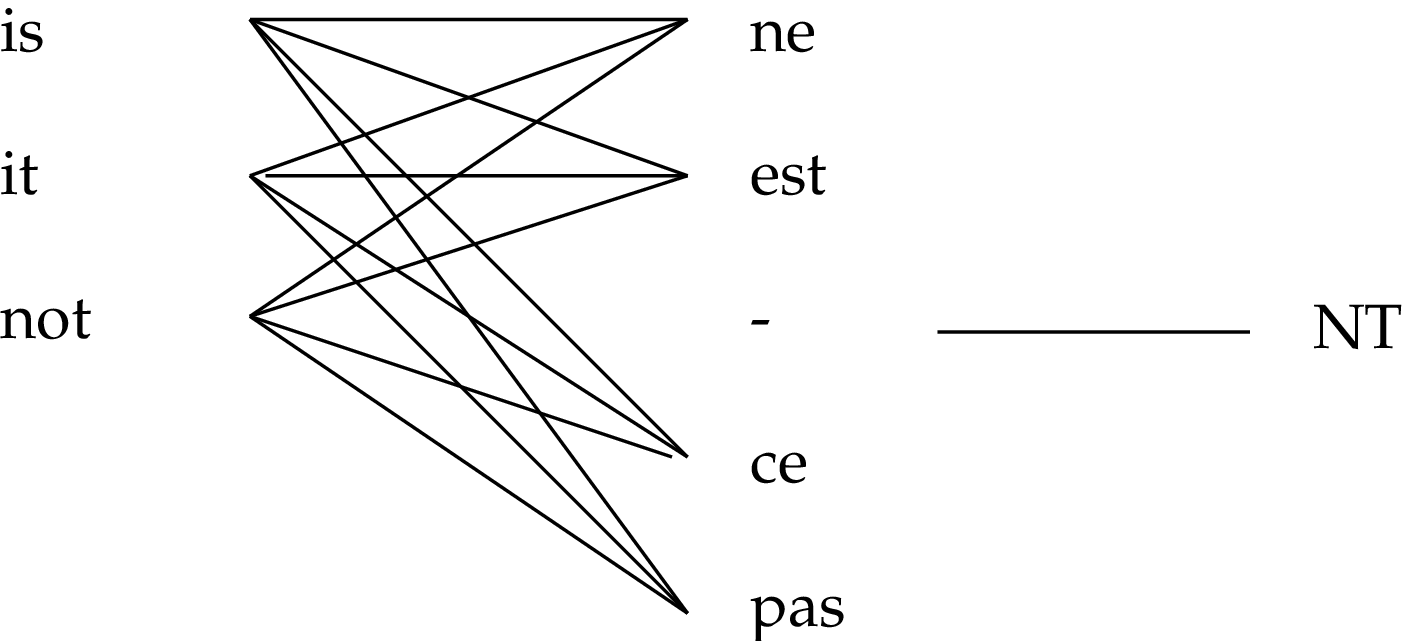,width=4.5in}}

\subsection{Idioms and Near Idioms}

``Frozen'' expressions that are unique to one language or the other
should be linked as wholes.  E.g.:

\vspace*{1in}

\centerline{\psfig{figure=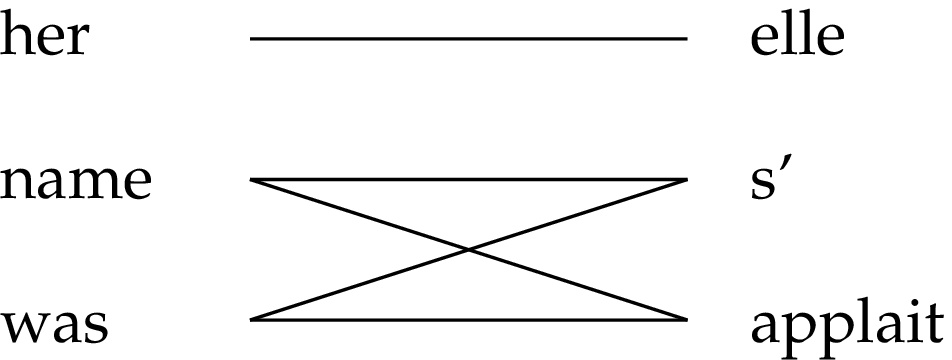,width=3in}}

\subsection{Referring Expressions}

\subsubsection{Pronouns and Definite Descriptions}

Divergent descriptions of the same thing should be linked as wholes,
as in Example~\ref{eg3}.  This rule holds even when one description is
a pronoun:

\vspace*{.3in}

\centerline{\psfig{figure=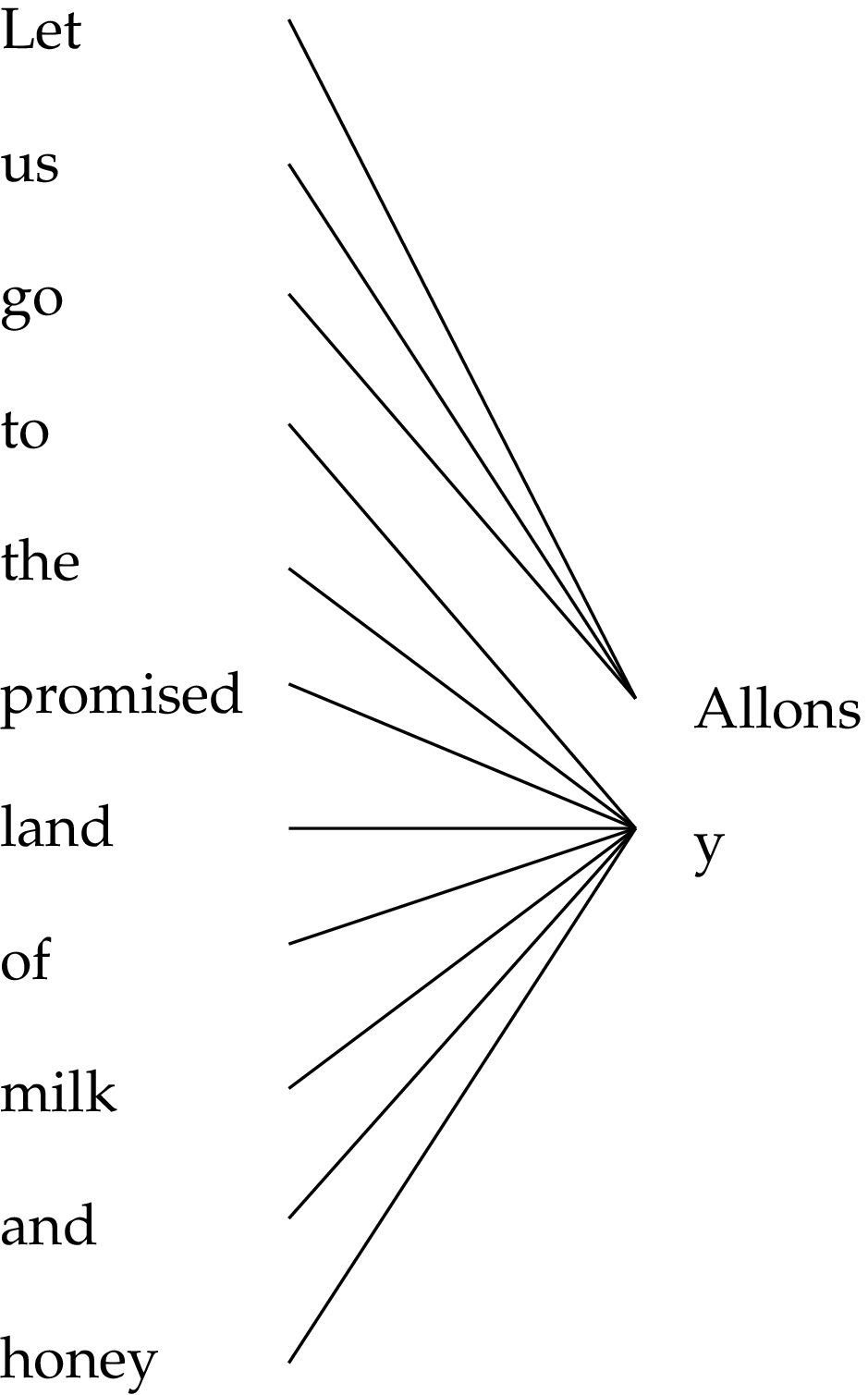,width=3in}}

\subsubsection{Resumptive Pronouns}

{\em Resumptive pronouns} refer to something previously described in
the same sentence, called the {\em antecedent}.  When a resumptive
pronoun occurs in a verse, but not in its translation, both the
resumptive pronoun and its antecedent should be linked to the
translation of the antecedent.  Relative markers should be treated the
same way.

\vspace*{.3in}

\centerline{\psfig{figure=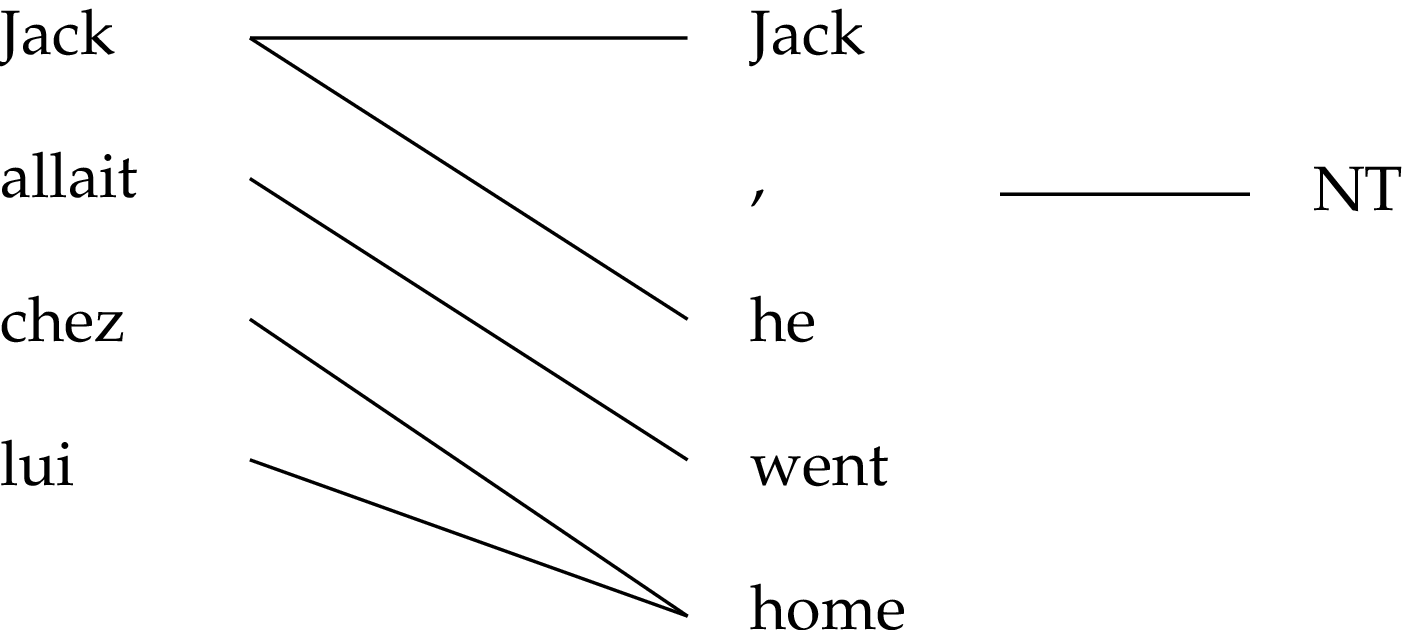,width=4in}}

\newpage

\centerline{\psfig{figure=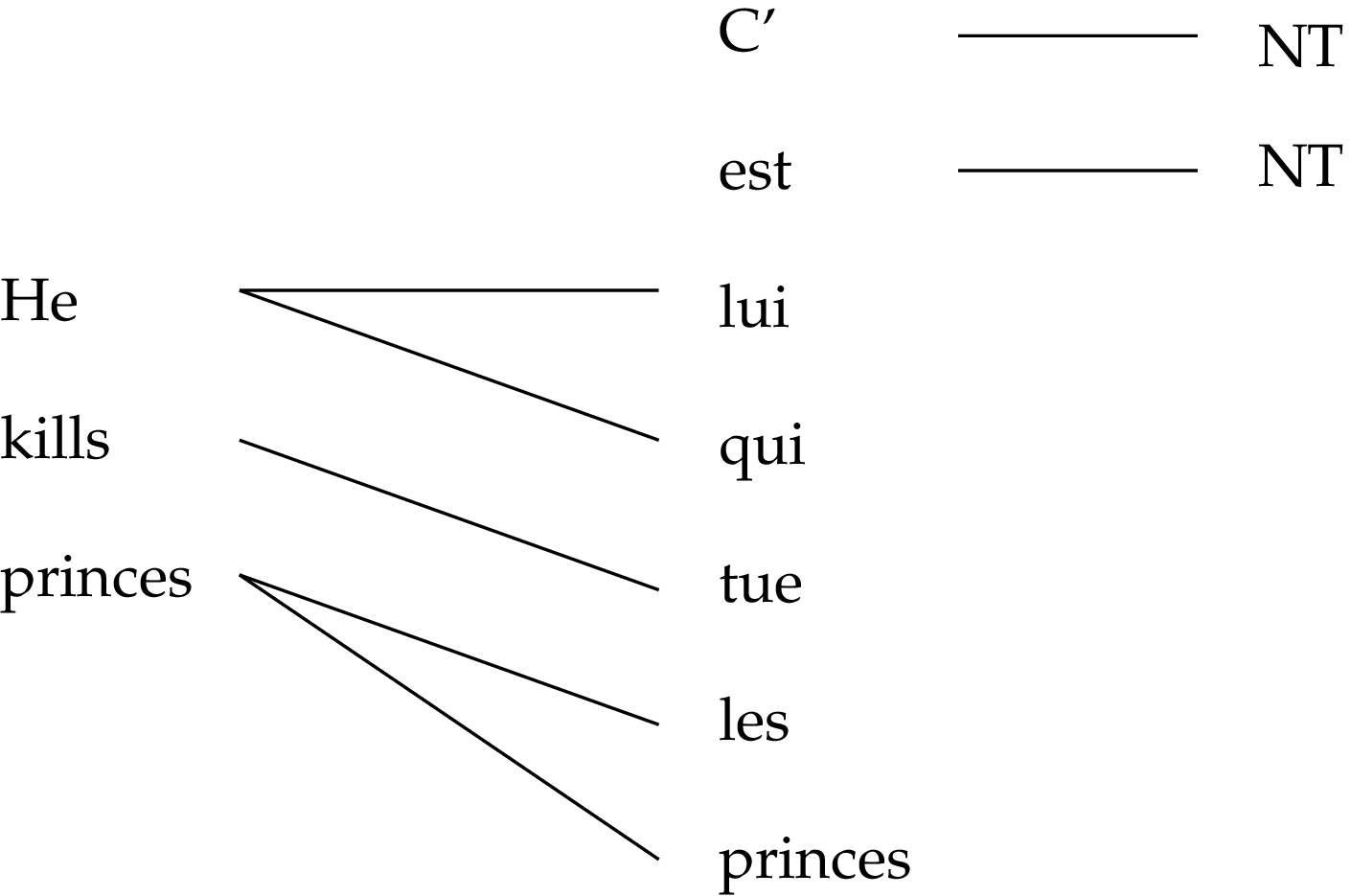,width=4.5in}}

\vspace*{1in}

\centerline{\psfig{figure=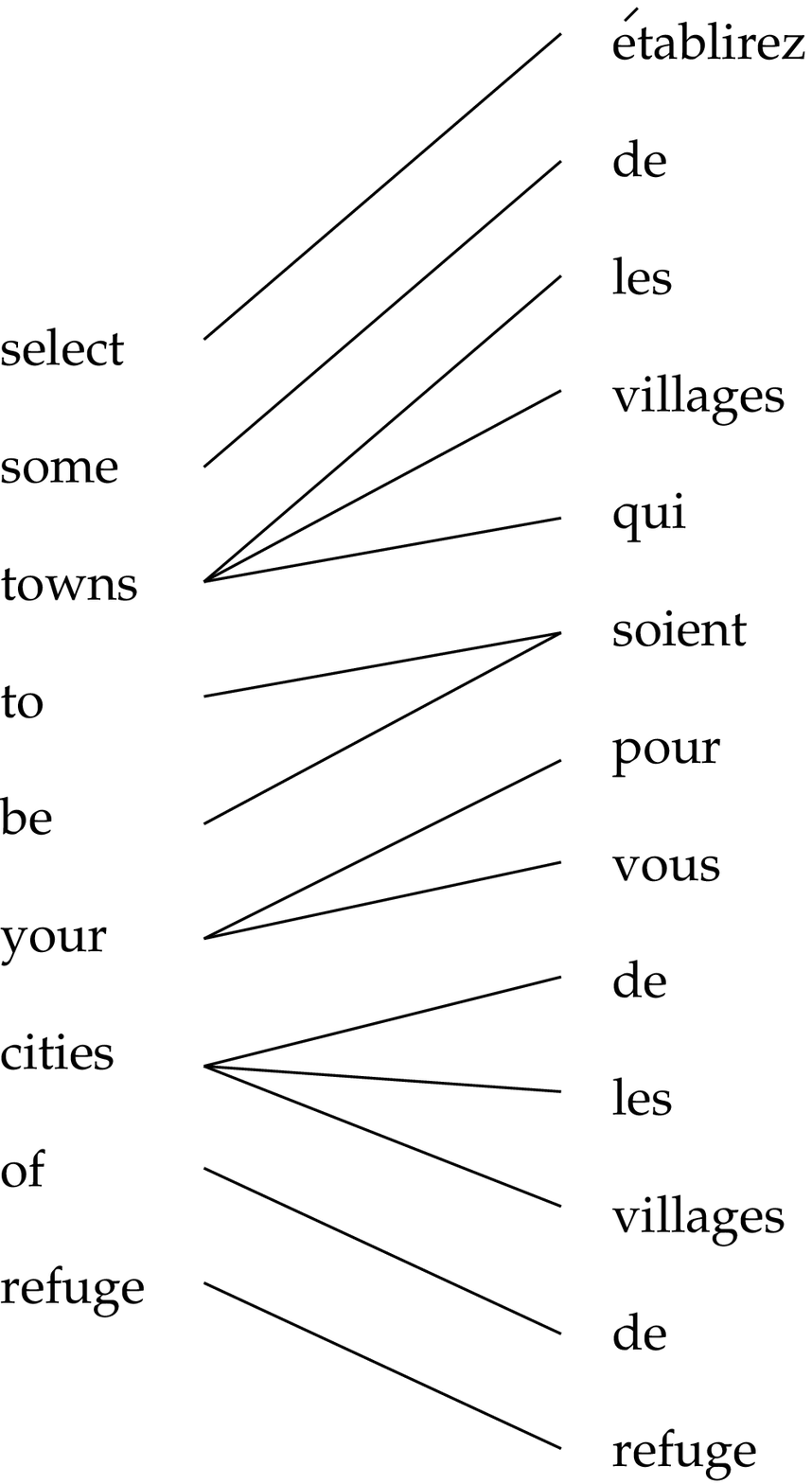,width=3in}}

\subsubsection{Conjunctive Non-Parallelism}

When a piece of text is repeated in a verse but not in its
translation, all instances of that piece of text in the first verse
should be linked to the one translation:

\vspace*{1in}

\centerline{\psfig{figure=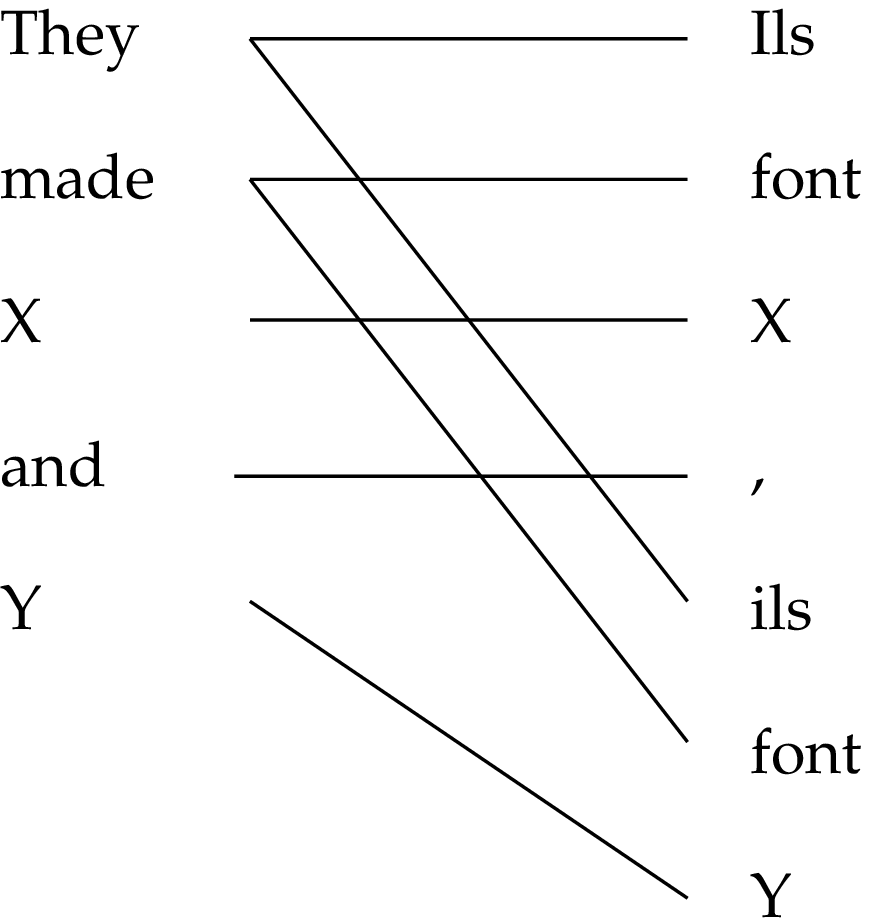,width=2.5in}}

\vspace*{1in}

\centerline{\psfig{figure=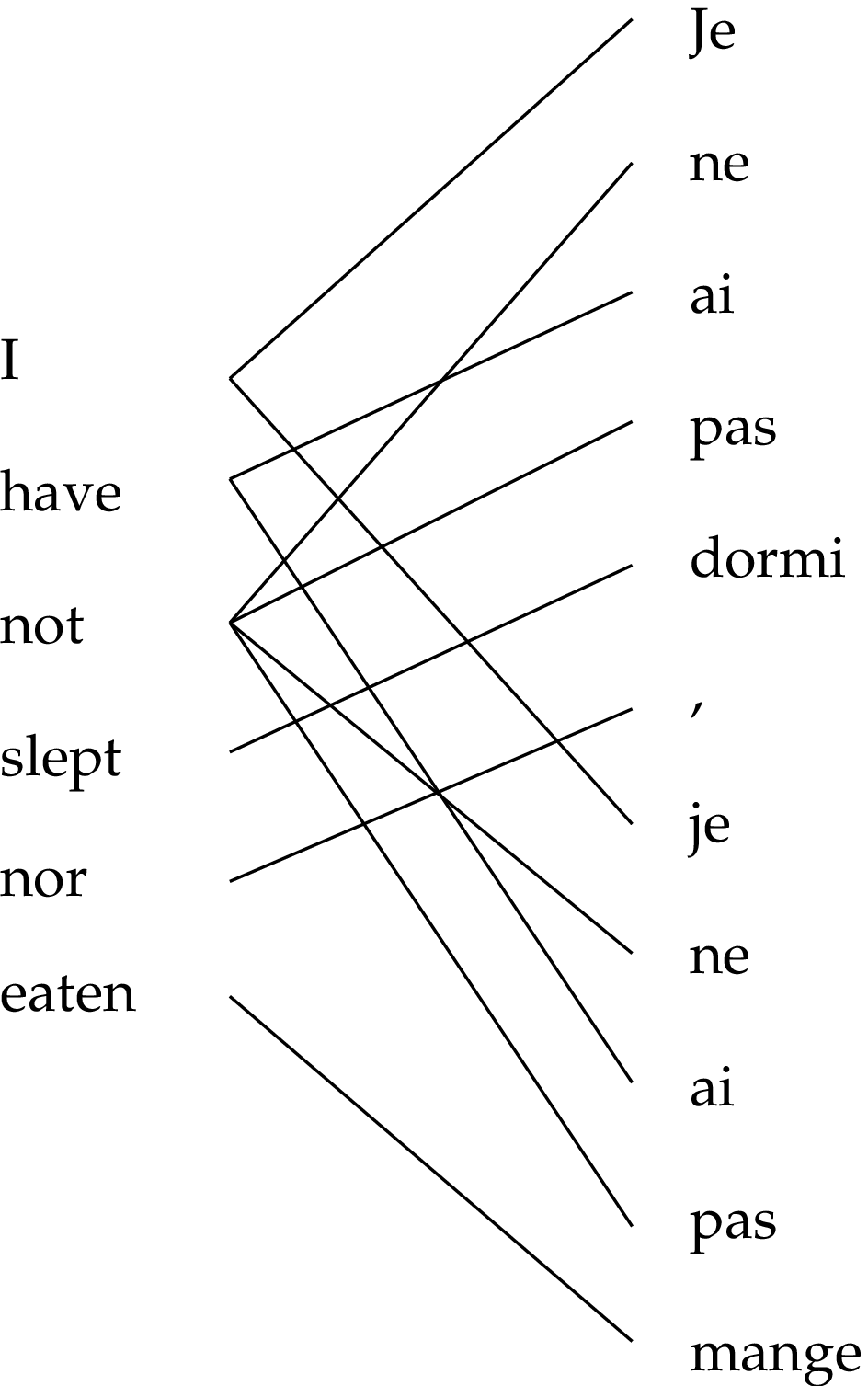,width=3in}}

\subsection{Verbs}

\subsubsection{Negation}

French negation often involves two words, where English uses
only one.  In all such cases, {\em both} pieces of the French
negation should be linked to the English negation.  Examples include
{\em ne \ldots pas, ne \ldots point, ne \ldots rien, ne \ldots jamais,
ne \ldots que}.

\subsubsection{Auxiliary Verbs}

Auxiliary verbs should {\em not} be linked to the main verb in the
translation whenever that main verb also has auxiliaries attached.
However, auxiliaries often do not match, especially when the verb
tenses get slightly altered in translation.  When there are
auxiliaries in one verse, but not in its translation, both the
auxiliaries and the main verb should be linked to the main verb in the
translation.  E.g.:

\vspace*{.5in}

\centerline{\psfig{figure=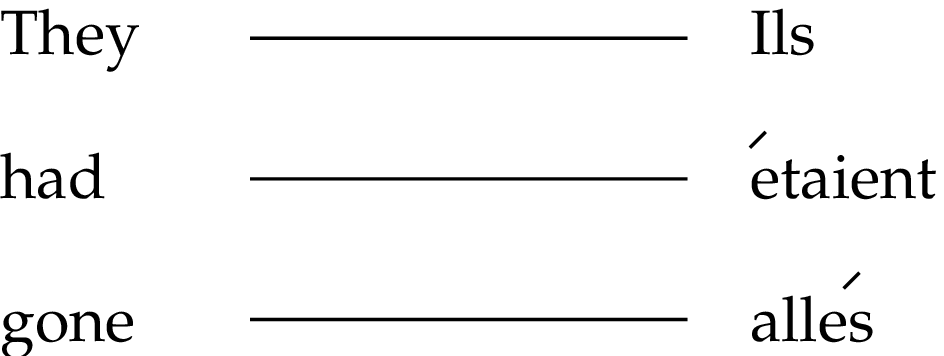,width=3in}}

\vspace*{.5in}

But consider {\em May \ldots be / soit}:

\vspace*{.5in}

\centerline{\psfig{figure=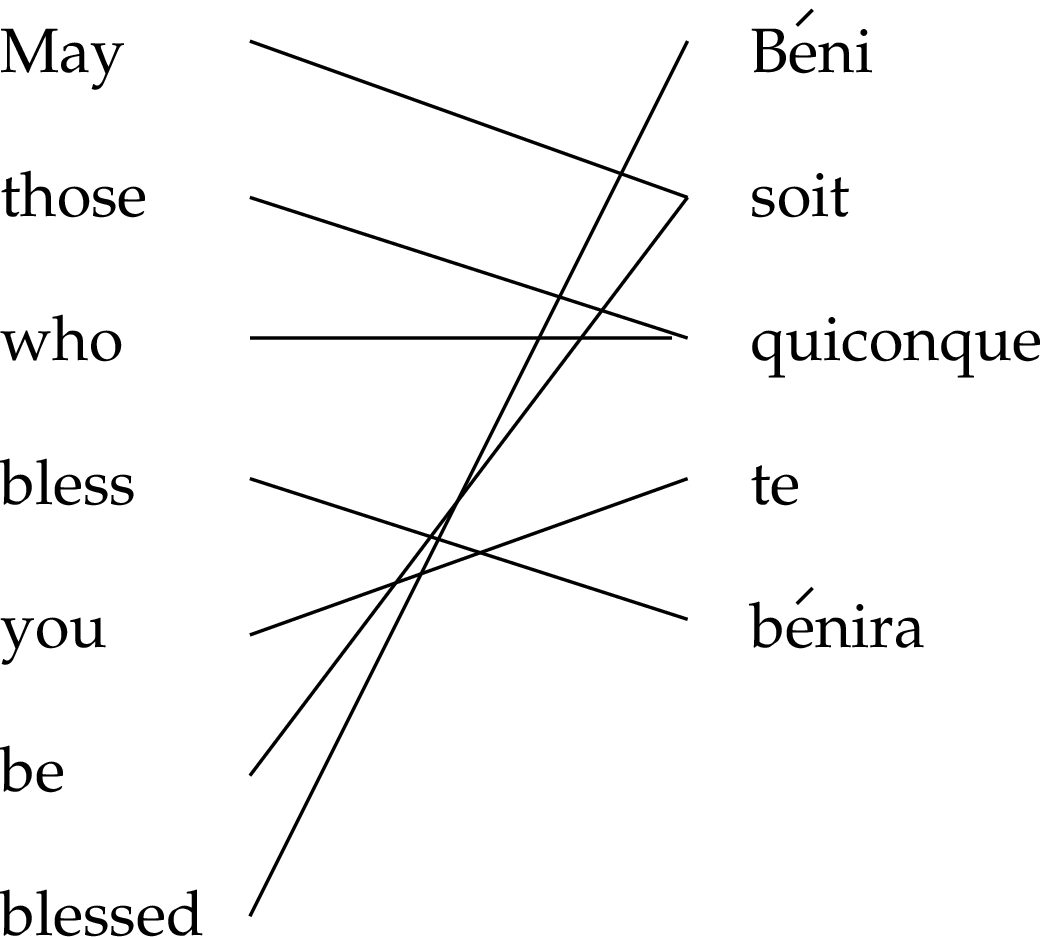,width=4in}}


\newpage

\centerline{\psfig{figure=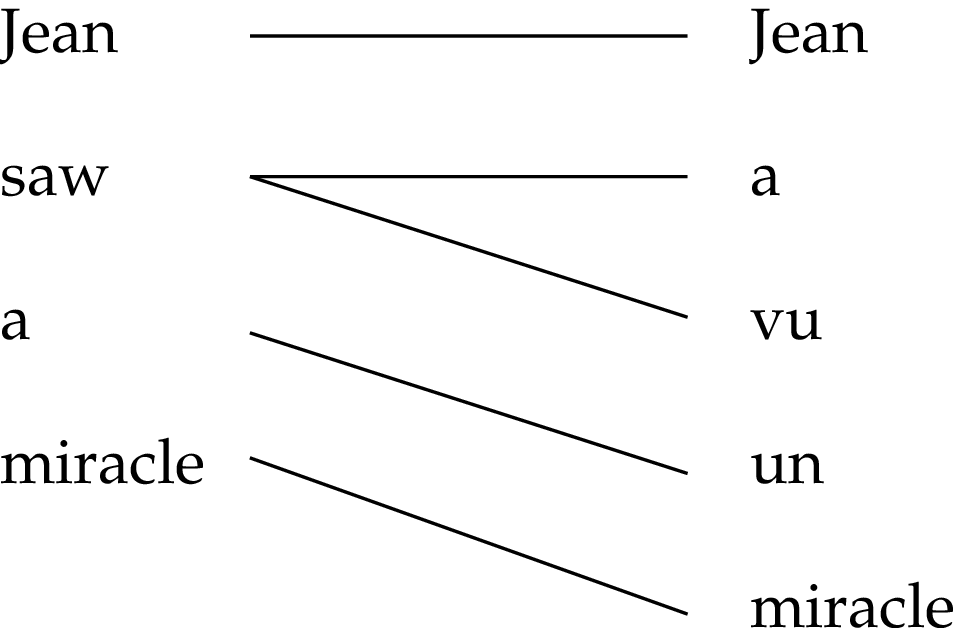,width=3in}}

\newpage

\subsubsection{Passivization}

The order of corresponding words in a pair of verses may be very
different when on verse is in the {\em passive voice} and the other is in
the {\em active voice}.  You should make an effort to tease apart the
correspondences, instead of linking whole phrases.  E.g.:

\vspace*{.5in}

\centerline{\psfig{figure=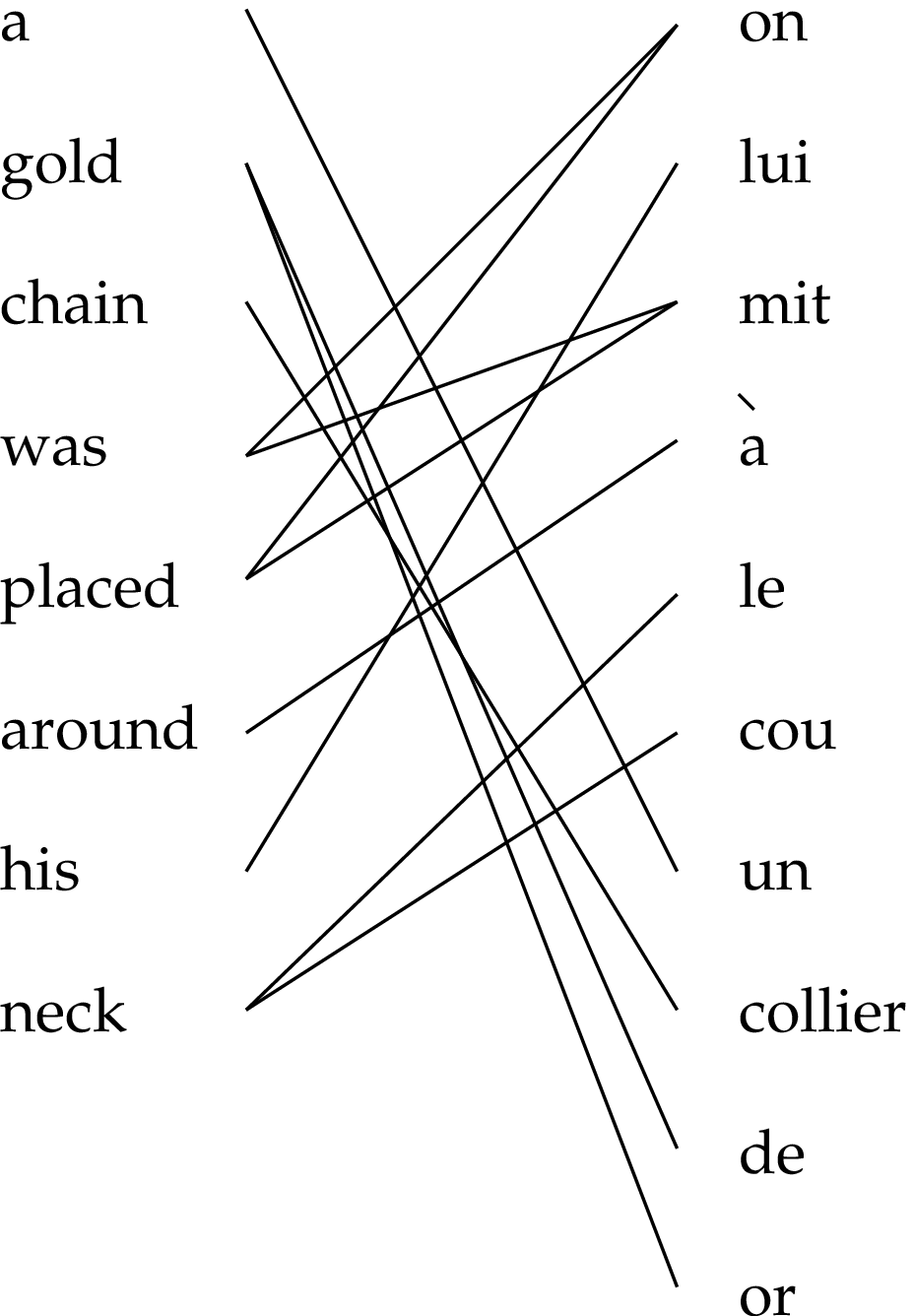,width=3in}}

\vspace*{.5in}


\centerline{\psfig{figure=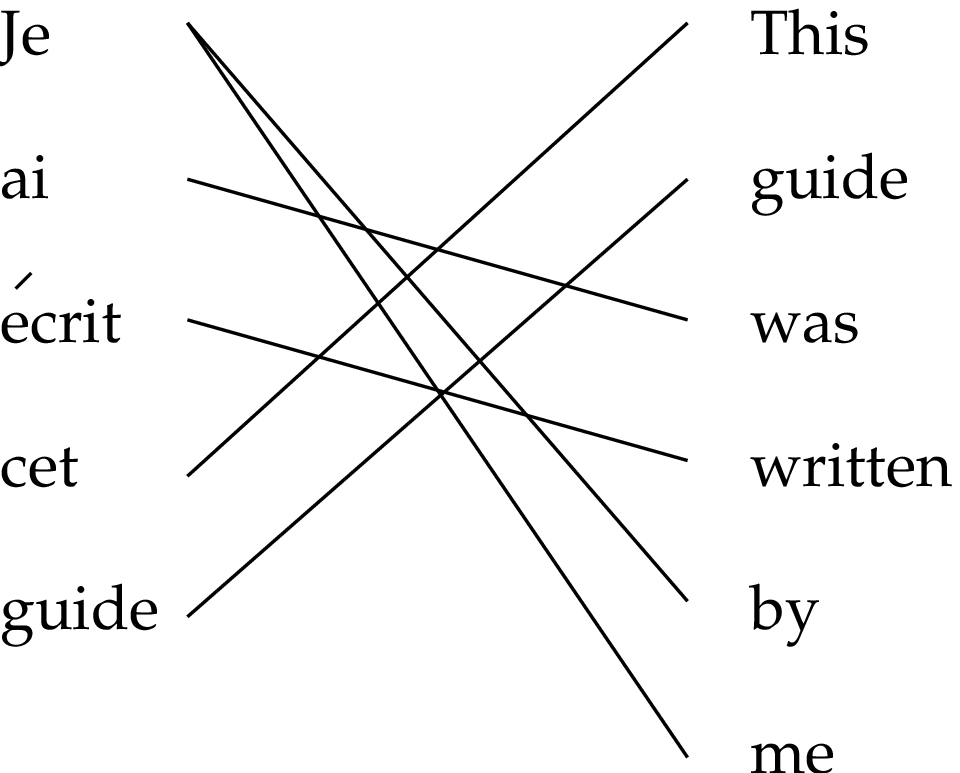,width=3in}}

\subsection{Prepositions}

\subsubsection{Extra Prepositions}

When a verse contains a preposition that does not appear in the
translation, the preposition should be linked to the translation of
the preposition's object, not the translation of its subject.  E.g.:

\vspace*{1in}

\centerline{\psfig{figure=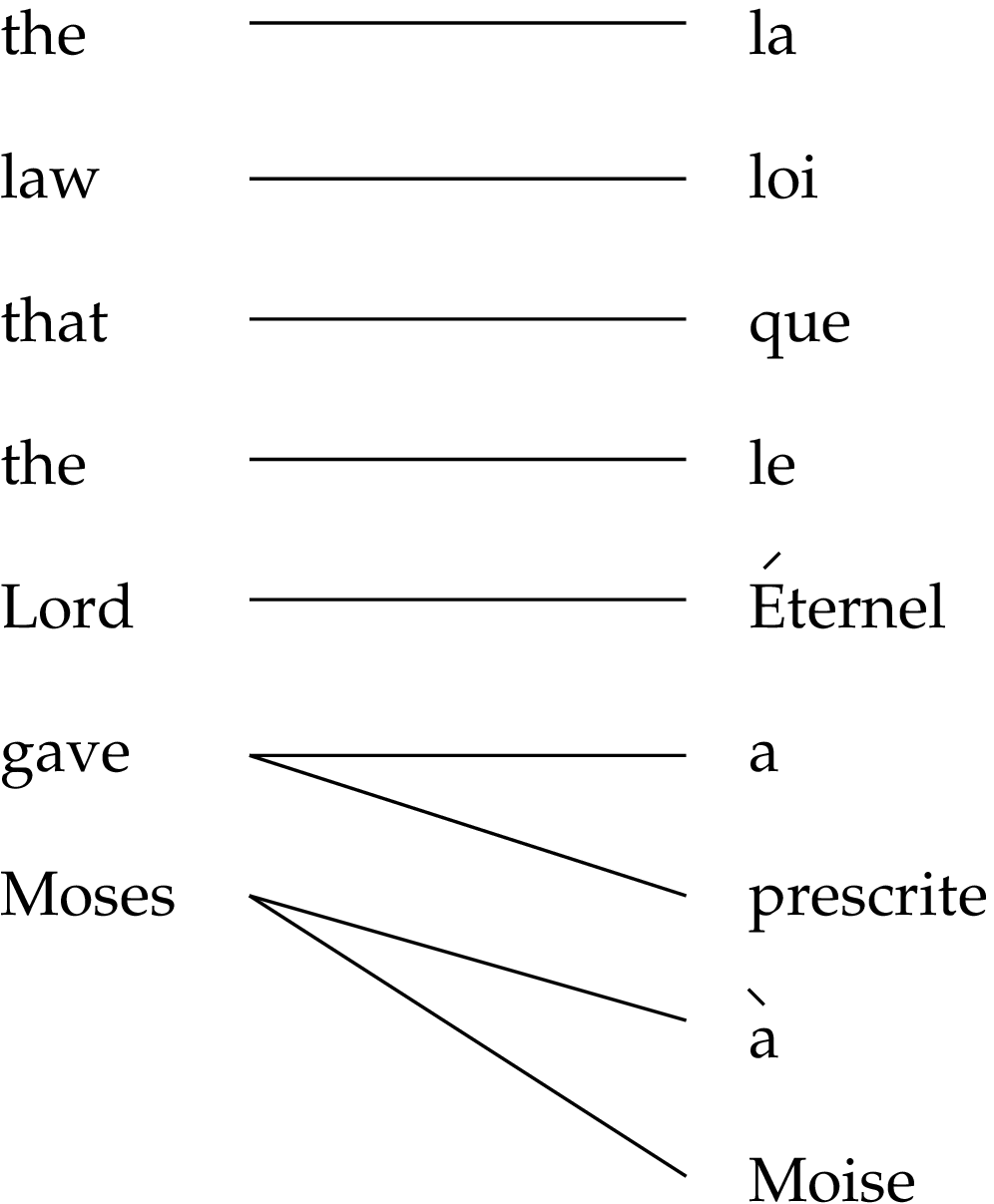,width=3in}}

\vspace*{1in}

\centerline{\psfig{figure=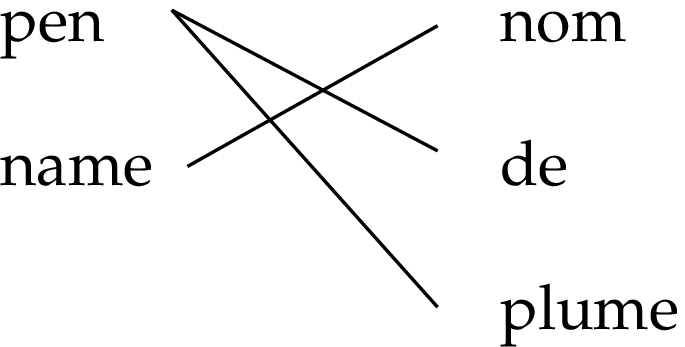,width=3in}}


\newpage

\subsubsection{Divergent Prepositions}

When a piece of text is slightly paraphrased, two prepositions that
never mean the same thing literally may need to be linked anyway:

\vspace*{1in}

\centerline{\psfig{figure=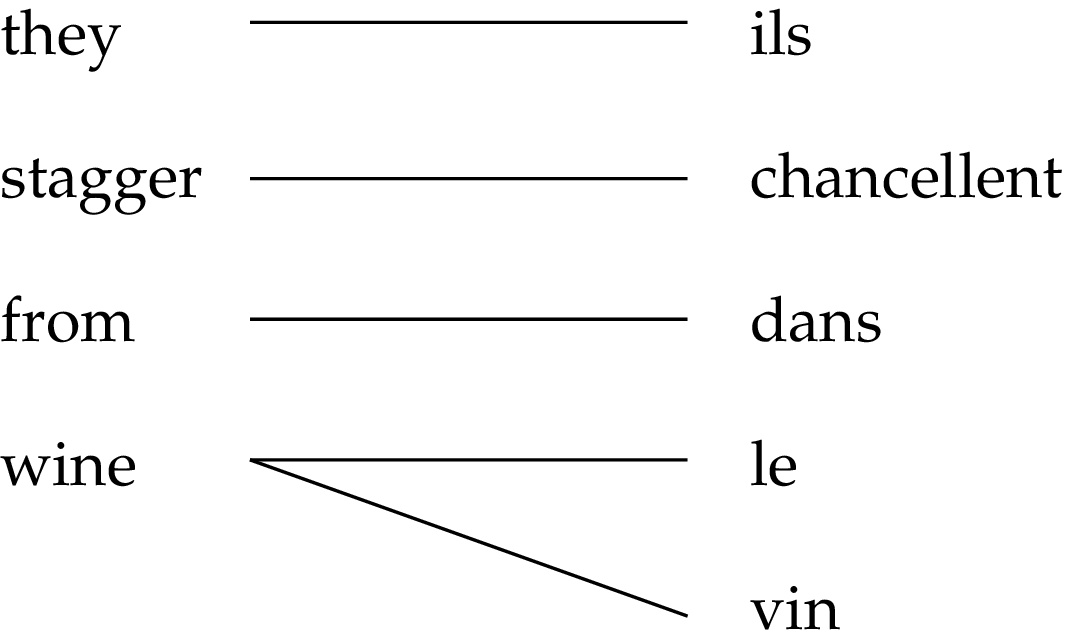,width=3in}}

\subsection{Determiners}

\subsubsection{Extra Determiners}

Extra determiners in a verse should be linked together with their noun
to the noun's translation:

\vspace*{1in}

\centerline{\psfig{figure=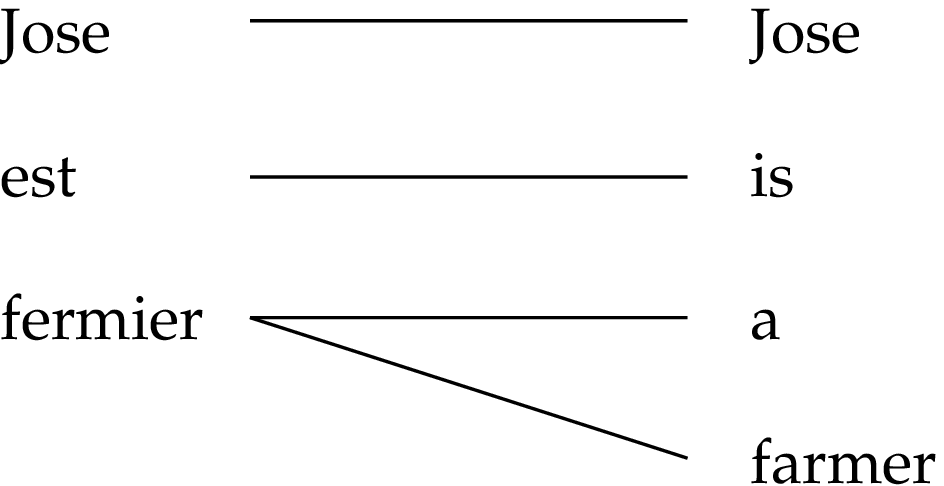,width=3in}}

\vspace*{1in}

\centerline{\psfig{figure=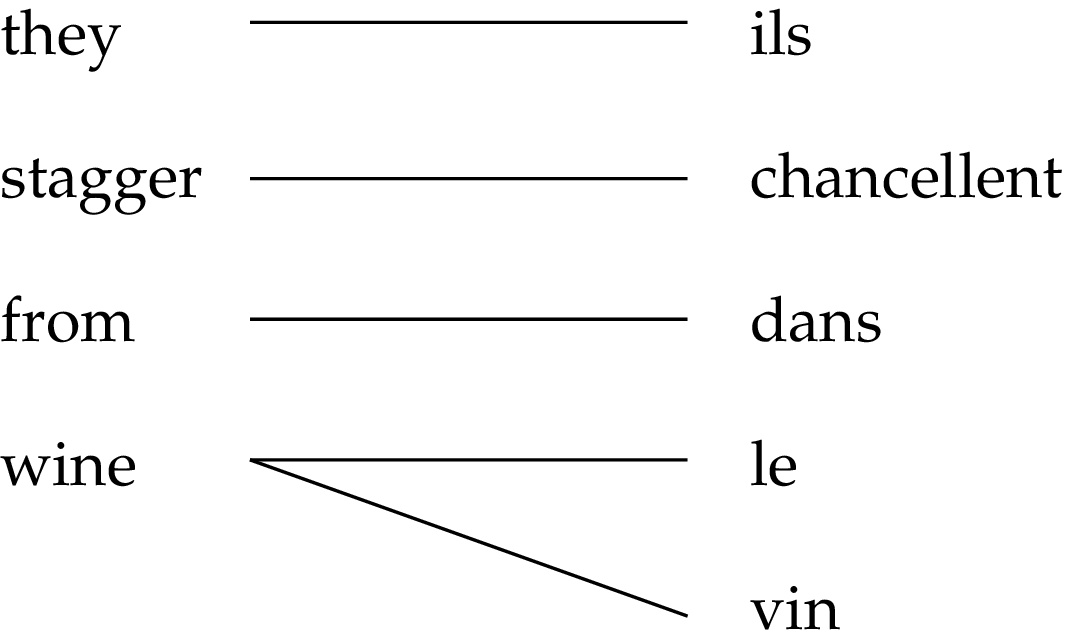,width=3in}}

\subsubsection{Possessives}

English and French possessive markers are different, but easy to
identify.  They should be linked separately from their nouns:

\vspace*{1in}

\centerline{\psfig{figure=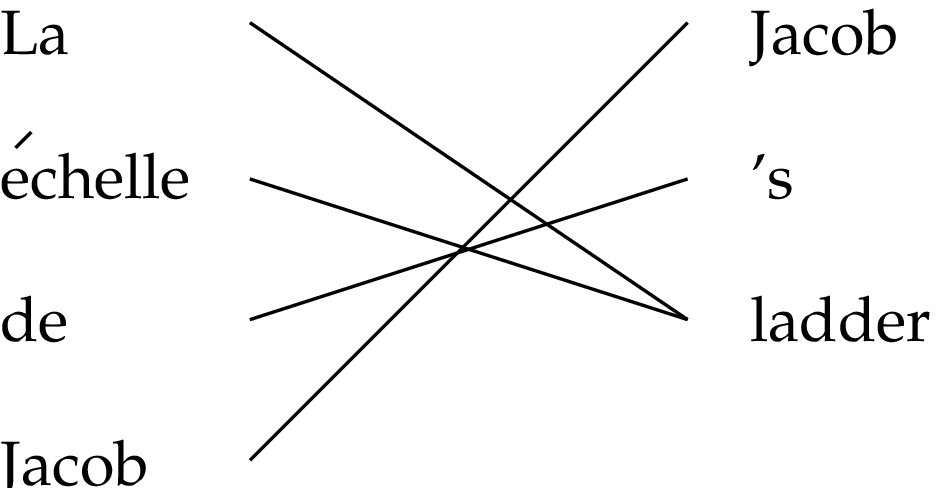,width=3in}}

\vspace*{1in}

The English plural possessive marker is just an apostrophe:

\vspace*{1in}

\centerline{\psfig{figure=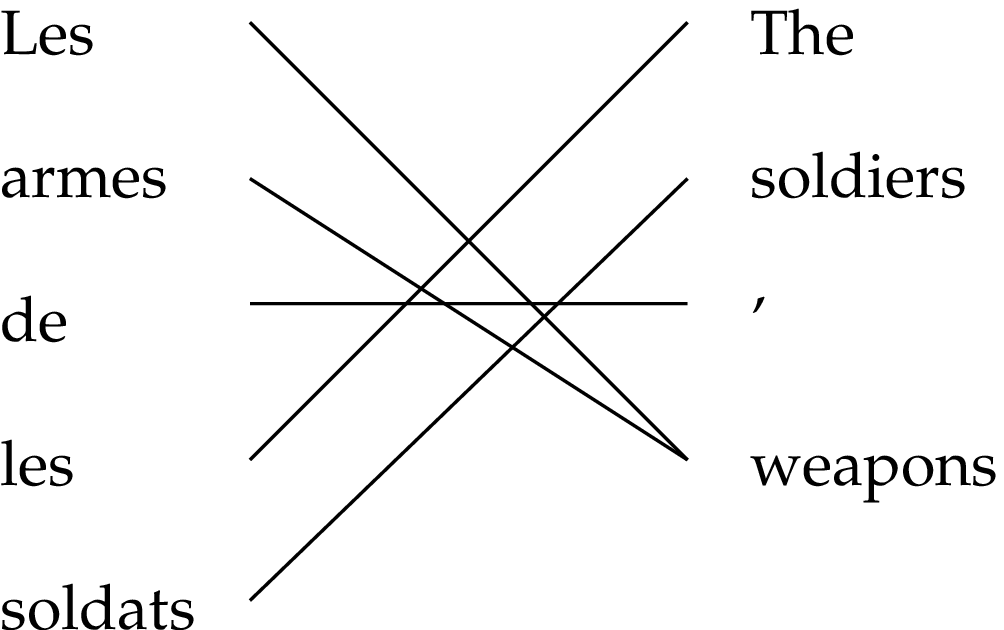,width=3in}}

\newpage

\subsection{Punctuation}

\subsubsection{Punctuation Series}

Sometimes a verse pair will contain several identical (or similar)
punctuation marks on each side, but in different quantities.  In such
cases, the best linking strategy is to link all the words other
than the punctuation marks first.  Then, link the punctuation marks to
minimize the number of ``crossing'' links.  E.g.:

\centerline{\psfig{figure=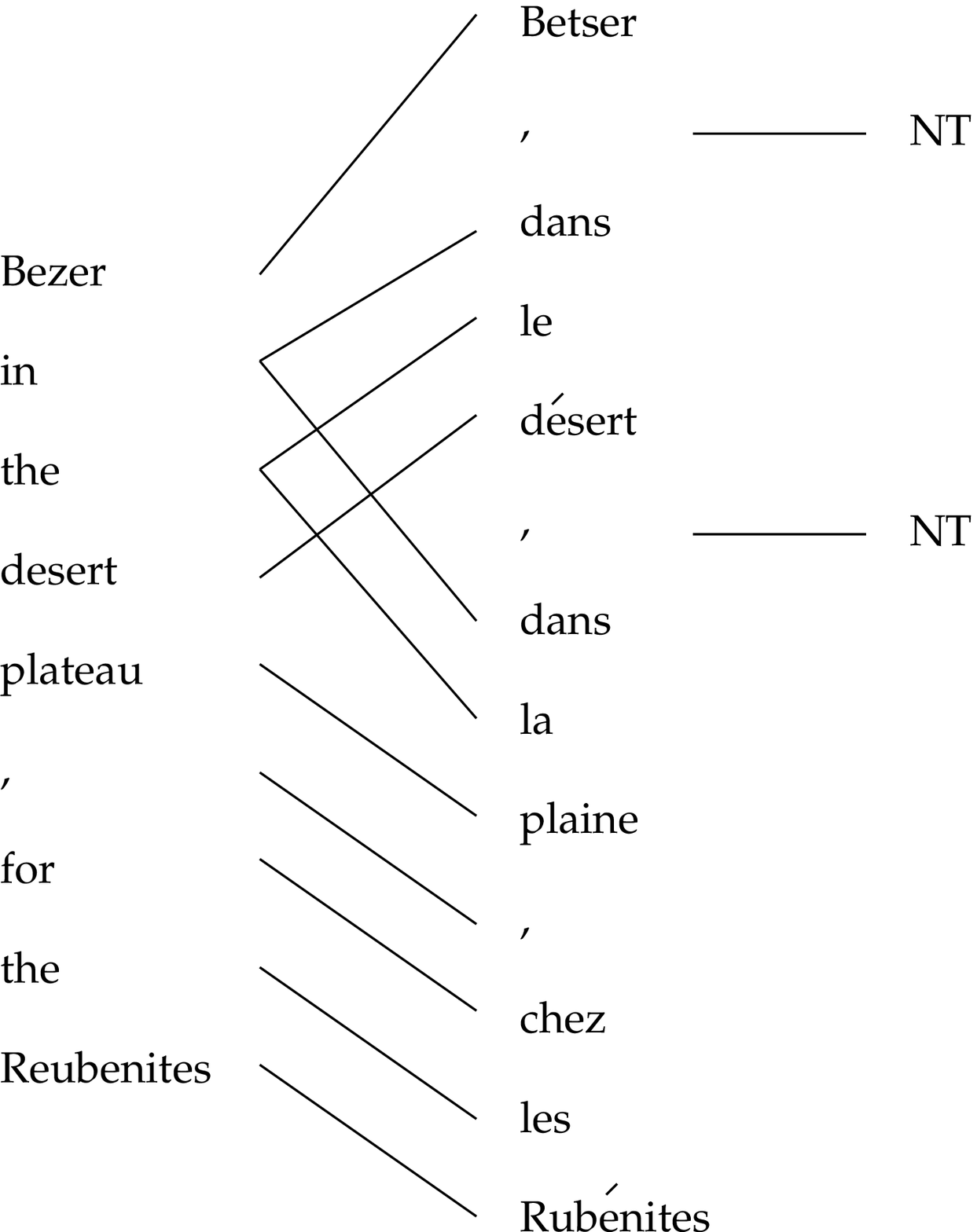,width=4.5in}}

\subsubsection{Punctuation and Conjunction}

When a series of conjunctions in one verse corresponds to a series of
punctuation marks in the other verse, don't hesitate to link word to
punctuation marks.  E.g, English ``and'' will often correspond to a
French comma.

\end{document}